\documentclass[10pt,twocolumn,letterpaper]{article}

\usepackage{wacv}
\usepackage{times}
\usepackage{epsfig}
\usepackage{graphicx}
\usepackage{amsmath}
\usepackage{amssymb}
\usepackage{authblk}
\usepackage{multirow}
\usepackage[accsupp]{axessibility}  

\def\wacvPaperID{****} 

\wacvfinalcopy 

\ifwacvfinal
\fi

\ifwacvfinal
\usepackage[breaklinks=true,bookmarks=false]{hyperref}
\else
\usepackage[pagebackref=true,breaklinks=true,colorlinks,bookmarks=false]{hyperref}
\fi

\begin{document}

\title{Predicting Levels of Household Electricity Consumption in Low-Access Settings}


\author[1]{Simone Fobi \thanks{We thank The Rockefeller Foundation for their generous support of this work under Grant 2018 POW 004.}}
\author[2]{Joel Mugyenyi}
\author[3]{Nathaniel J. Williams}
\author[1]{Vijay Modi}
\author[2]{Jay Taneja}
\affil[1]{Department of Mechanical Engineering, Columbia University}
\affil[2]{Department of Electrical \& Computer Engineering, University of Massachusetts Amherst}
\affil[3]{Golisano Institute for Sustainability, Rochester Institute of Technology}
\affil[ ]{\tt\small \{sf2786,modi\}@columbia.edu, \{jmugyenyi,jtaneja\}@umass, njwgis@rit.edu }

\maketitle

\ifwacvfinal
\thispagestyle{empty}
\fi

\begin{abstract}
In low-income settings, the most critical piece of information for electric utilities is the anticipated consumption of a customer. Electricity consumption assessment is difficult to do in settings where a significant fraction of households do not yet have an electricity connection. In such settings the absolute levels of anticipated consumption can range from 5-100 kWh/month, leading to high variability amongst these customers. Precious resources are at stake if a significant fraction of low consumers are connected over those with higher consumption.
This is the first study of it's kind in low-income settings that attempts to predict a building's consumption and not that of an aggregate administrative area.  We train a Convolutional Neural Network (CNN) over pre-electrification daytime satellite imagery with a sample of utility bills from 20,000 geo-referenced electricity customers in Kenya (0.01\% of Kenya's residential customers). This is made possible with a two-stage approach that uses a novel building segmentation approach to leverage much larger volumes of no-cost satellite imagery to make the most of scarce and expensive customer data. Our method shows that competitive accuracies can be achieved at the building level, addressing the challenge of consumption variability. This work shows that the building's characteristics and it's surrounding context are both important in predicting consumption levels. We also evaluate the addition of lower resolution geospatial datasets into the training process, including nighttime lights and census-derived data. The results are already helping inform site selection and distribution-level planning, through granular predictions at the level of individual structures in Kenya and there is no reason this cannot be extended to other countries.
\end{abstract}

\section{Introduction}
Improved engineering and new business models for electrification have contributed to increasing access to electricity around the world. However, 840 million people still lack access to electricity services \cite{WB}, many of them residing in places that are difficult to reach and, as a result, expensive to serve~\cite{zvoleff2009impact, parshall2009national}. Energy providers, constrained by limited investment budgets, face a perpetual trade-off between expanding electricity access and cost recovery. When consumption levels are low, as can occur in low-income settings, utilities struggle to recover the cost of servicing a grid connection, and the government subsidies\cite{fobi2018longitudinal} for initial capital are poorly utilized. Alternatives to grid extension such as Solar Home Systems (SHS) can support smaller loads without the large wire investments, while in some cases clustered homes (with clusters far from each other) can make mini-grids viable\cite{fobi2021scalable}. In practice, identifying those likely to become high consumers is critical to the energy provider, as these are critical to revenue generation and system cost recovery. Given the diversity of electrification technologies, planners rely on energy access planning tools (e.g., the \textit{Open Source Spatial Electrification Tool (OnSSET)}\cite{onsset}) that utilize electricity consumption tiers, to match potential customers with technologies that can cost-effectively meet consumption. Consumption predictions can  assist matching areas with cost-effective energy technologies, enabling a country to provide electricity access to a larger population given the same investment.

We make four unique contributions. \textit{First}, we introduce a data-driven method to predict levels of future electricity consumption for individual households, using information prior to the household being electrified. Our approach trains a CNN to predict levels of household consumption using pre-electrificaion daytime satellite images. Although accurate individual household electricity consumption predictions are difficult to achieve~\cite{streltsov2020estimating}, we show that high-resolution daytime satellite imagery (0.5 m/pixel) performs better (preserving performance at different levels of consumption) than other approaches (historical consumption, census indicators, and Nighttime Lights) that result in heavily-skewed prediction performance.
\textit{Secondly}, our proposed method shows that learning about buildings through a building segmentation task and over a large volume of images improves the downstream task of electricity consumption prediction.
\textit{Thirdly}, we demonstrate a method for model interpretation that quantifies the importance of building characteristics relative to the surrounding context. Specifically, we show that building roof sizes and color are relevant to predicting consumption levels. Our approach also shows that learning about the household's surrounding context improves prediction performance between 2-5\% depending on the consumption tier.
\textit{Finally}, we present additional validation of our results using the World Bank's Multi-Tier Framework survey of electricity consumption among households. Sample weighted Pearson correlation scores between the survey and our predictions for 5.3 million residential buildings were 0.82 when excluding the over-sampled and already-electrified capital city of Nairobi, and 0.64 otherwise. Outputs from our model can be used in planning tools such as OnSSET, which utilize electricity demand tiers as an input parameter for spatial electrification planning. Given the potential dependence of consumption on tariffs and policies for recovery of installation costs, the specific results of this data-driven approach apply to Kenya. However, our methods can be extended to other countries, thereby offering insights to electricity planners.
\section{Related Work}
\textbf{Predicting Electricity consumption}:
\cite{SWAN20091819, BHATTACHARYYA20101979} present a comprehensive survey on residential energy consumption prediction. First, we review load forecasting in time for individual households, highlighting how the problem at hand is different. One approach \cite{gajowniczek2017electricity} forecasts individual household level electricity loads 24 hours ahead using sequence mining and smart meter data. Another \cite{ushakova2020big} clusters customer smart meter data into behavioral groups and later use supervised techniques such as Random Forest to predict customer clusters given unseen smart meter data. Other residential load forecasting studies \cite{sajjad2020novel,hong2020deep,kong2017short} also predict the short-term future consumption of houses given their historical data or appliance usage, and deep learning techniques. Past consumption data is essential for such studies and hence not suitable for future consumption prediction where no prior data exists. Some studies have attempted to address the problem of predicting the future consumption of a currently unelectrified household. In \cite{syed2013estimation} the average consumption of previously connected customers (by municipality) is used to estimate consumption for unelectrified households. This would not capture the variations amongst households. \cite{shen2017household} use support vector regression to study the relevance of 48 household survey variables (demographics, appliance ownership, household personality traits) in predicting household consumption. \cite{pandyaswargo2020estimating} use an energy end-use model to estimate demand for off-grid communities in Myanmar, Indonesia and Laos through household surveys that measure appliance ownership and usage. \cite{olaniyan2018estimating} take a similar approach to estimating residential electricity consumption in Nigeria by collecting survey responses on appliance ownership and usage. \cite{allee2021predicting} use machine learning to predict daily electricity consumption tiers upon connecting to a microgrid, using features obtained from customer application surveys pre-electrification.  All of these studies use data that would be difficult and/or expensive to obtain at scale for a country. Our approach provides a scalable and faster approach to estimating consumption from proxy household features which are available in satellite imagery. 

\textbf{Satellite Imagery and Machine Learning}:
Recently, there has been a surge of studies applying CNNs to satellite imagery to assess building damage \cite{shen2021bdanet}, measure road quality \cite{cadamuro2019street}, estimate population density\cite{doupepop}, detect solar farms \cite{hou2019solarnet}, segment roads and buildings\cite{Demir_2018_CVPR_Workshops}, estimate rooftop density by type \cite{varshney2015targeting} and measure poverty. One approach \cite{jean2016combining} predicts wealth for multiple African countries by combining overhead daytime images with CNNs. The authors use high resolution daytime images in training a CNN to predict nighttime lights; features extracted from the trained model were then used to estimate household expenditure and wealth at a 10 x 10 km resolution. Their results suggest that predictions about economic development can be made from satellite image derived features; this insight provides additional motivation for developing methods that extract information from imagery for electricity consumption prediction. Building on \cite{jean2016combining}, multiple works \cite{head2017can, yeh2020using,das2020, tingzon, ledesma2020interpretable,ayush2021efficient,han2020learning} have assessed wealth, poverty and development using satellite images. \cite{falchetta2019high} use VIIRS nighttime lights, gridded population data and land cover to estimate binary electricity access rates and electricity consumption tiers at 1 x 1 km grids. These studies demonstrate the value of satellite imagery in serving as a proxy measure for varying features such as poverty, electricity access and consumption. However, all these studies are carried out at a larger spatial scale to preclude evaluations of poverty levels or electricity consumption tiers of individual households\footnote{Household consumption tiers are relevant for electrification planning}.
\cite{streltsov2020estimating} is the only study to the best of our knowledge that predicts individual building energy consumption using overhead imagery in Gainesville, Florida, and San Diego, California. While performance improves at a spatially aggregated level, at the individual building level, they report low correlation (r$^2$=0) between predictions and the training data in Gainesville. The context of this study is quite different given the consumption levels, the size and formal construction practices in the U.S.. Also, the problem they address is that of estimation for electrified households rather than consumption prediction for unelectrified households. 

To the best of our knowledge, ours is the first study of its kind that predicts electricity consumption at an individual household level using overhead imagery. We formulate our task as a classification rather than a regression problem, and provide model interpretation around learned features.

\section{Models}
\label{models}
\subsection{Problem definition}
Given a set of households found in buildings \textbf{B} = $\{b_1, b_2, b_3 ...b_n\}$, where each building has a corresponding satellite image prior to the household being electrified \textbf{X} = $\{x_1, x_2, x_3 ...x_n\}$, the objective of our proposed model $\mathcal{F}$($x_i$) is to use each building's corresponding satellite image pre-electrification, to predict its consumption class ($\hat{y}$) after the building has been electrified (i.e. $\hat{y_i}$= $\mathcal{F}$($x_i$)). Data from electric meters after electrification serve as ground truth values. Binary labels y$_i$ are obtained by applying a threshold \textit{thres} (e.g. $<=$ \textit{thres} kWh/month) to the average monthly consumption values of each individual household given its electric meter readings.

\begin{figure*}[h]
  \centering
  \includegraphics[width=13cm, height=8cm]{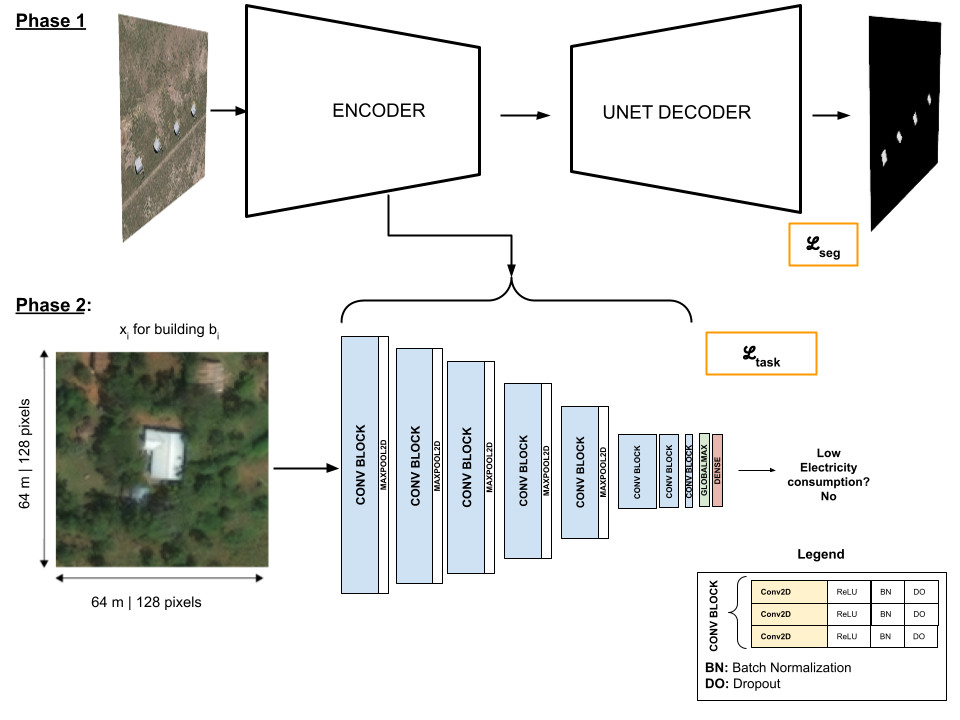}
  \caption{Phase 1: A building segmentation model is trained using the encoder in Phase 2 and a UNET decoder. The segmentation model is trained with a dissimilarity loss ($\mathcal{L}_{seg}$). Skip connections are omitted to maximize information funneling. Phase 2: The pretrained encoder is used in phase 2 to learn the electricity prediction task. An image ($x_i$) containing a household's building is input into the pretrained encoder. This encoder is trained with the negative log-likelihood ($\mathcal{L}_{task}$) loss to predict electricity consumption levels upon electrification.}
  \label{fig:arch}
  \vspace{-0.1in}
\end{figure*}

We propose a two-phase supervised method to predict binary consumption classes ($\hat{y_i}$). First, we prioritize learning about building features through the help of a building segmentation task. Next, the building segmentation model is used to initialize a supervised model to classify consumption levels. Figure \ref{fig:arch} illustrates the steps used for training, and their corresponding losses.
Our method is compared to other models that match commonly used approaches to predict electricity consumption levels after electrification. 
\subsection{Electricity Prediction Models}
Here we present 4 different approaches to predicting electricity consumption. The first two are widely used approaches, the third is our proposed method, and the fourth is an approach to support the interpretation of our method.
\subsubsection{Model A: Average Historical Consumption}
\textit{Model A} represents the simplest model and serves as a baseline. This method mimics approaches commonly used by energy planners to estimate consumption for the unelectrified. Energy providers have historical consumption data for already-connected households. Thus, this model assumes that the average  historical consumption of households in the same administrative unit is sufficient to approximate the consumption of customers that will be electrified in the future. In the case of Kenya, energy planning is done at administrative levels (following Kenya's policy of decentralized energy planning).
For each connection year $t_k$, households electrified prior to $t_k$ are used to calculate the average monthly consumption ($\bar{y}(c_j,t_k)$) of each constituency $c_j$. A constituency is an administrative unit and there are 290 constituencies across the country. Households who share the same constituency and electrification year are assigned the same $\bar{y}(c_j,t_k)$ value. A threshold value \textit{thres} (e.g. $<=$ \textit{thres} kWh/month) is applied to the assigned consumption of every household to determine the expected consumption class ($\hat{y_i}$). $\hat{y_i}$ is compared to the true consumption class $y_i$.
\subsubsection{Model B: MLP with Non-Visual Data}
Varying lower-resolution datasets are widely available and can serve as proxies for electricity access. In fact \cite{falchetta2019high,han2020learning} use non-visual data to evaluate electricity access and economic development. We present \textit{model B}, based on publicly available non-visual datasets to evaluate their performance in predicting consumption levels upon electrification. \textit{Model B} offers more complexity than an average historical consumption strategy. Non-visual features are inputted into a Multi-Layer Perceptron (MLP) containing 3 dense layers with 64, 32, and 16 filters respectively (Supplementary Material: A.1.). The model is trained by minimizing the Negative Log-Likelihood loss as shown in Equation \ref{bce_loss}.
\begin{equation}
\label{bce_loss}
  \mathcal{L}_{task} = \sum_y {-\log(p(y_i;\theta))} , 
\end{equation}
$\theta$ and y$_i$ are the model weights and consumption labels.

\subsubsection{Model C: Building Characteristics and Context}
\textit{Model C} combines both information about building characteristics with information about the surrounding context. In this model both the building of interest ($b_i$) and its surrounding context (in the form of a 128 x 128 image patch pre-electrification) are used to predict consumption levels post-electrification. Electricity consumption levels post-electrification are learnt in two phases. First we train an encoder-decoder building segmentation model to learn relevant building features. Next, we extract the trained encoder, add a classifier head and use the learnt building weights to initialize the consumption prediction task. Below we present a description of each phase as shown in Figure \ref{fig:arch}.

\textbf{Phase 1: Learning about buildings}: Deep learning has been shown to thrive in the presence of large amounts of labels. Although our electricity billing dataset is the largest of its kind (i.e., in a similar context) ever studied, its size remains small relative to the amounts frequently used to train data-hungry CNNs. We hypothesize that learning a proxy task (such as building segmentation) could provide relevant image encodings for predicting levels of electricity consumption, especially when small numbers of labels are available. We employ a much larger dataset of 6,928,078 building footprint polygon geometries in Uganda released by Microsoft\cite{microsoftug} for building segmentation. Building polygons from Uganda are used because there is no large high quality building footprint data in Kenya, and Uganda is the closest geographic country to Kenya with building polygons.\footnote{This work was done prior to the release of Google Footprints \cite{google_buildings}} Noisy (misaligned or missing) building polygons were observed within the Microsoft data in some parts of Uganda. 
Nevertheless, RGB patches of 128 x 128 pixels were used to train the building segmentation model in Uganda. We combine a custom encoder with a UNET-decoder to perform building segmentation (Figure \ref{fig:arch}). This encoder architecture is used both as an encoder for building segmentation and as an encoder for the classifier in phase 2. This architecture was inspired by the DeepSense architecture\cite{unet_deepsense} and has been shown to be helpful in remote sensing applications such as building segmentation. Skip connections between the encoder-decoder are excluded to maximize information funnelling through the encoder during phase 2. 64 filters were used in each layer of the encoder. The building segmentation model was trained with a dissimilarity loss ($\mathcal{L}_{seg}$) as shown in Equation \ref{jaccard_loss}, which builds on the Jaccard index $\mathcal{J}$(U,$\hat{U}$).
\begin{equation}
\label{jaccard_loss}
    \mathcal{L}_{seg} = 1 - \mathcal{J}(U,\hat{U}) = 1 - \frac{ (U \cdot \hat{U}) + \epsilon }{(U + \hat{U} - U \cdot \hat{U}) + \epsilon}
\end{equation}

where U represents the true footprints, $\hat{U}$ represents the predicted footprints and $\epsilon$ is used for numerical stability. The learnt encodings are later used in the downstream consumption level prediction task to bootstrap the classifier.

\textbf{Phase 2: Predicting electricity consumption levels}: After training the building segmentation model using Uganda data, the encoder-decoder network is initialized with the best building segmentation weights. The encoder is extracted and merged with a classification head (consisting of a global max-pooling and a dense layer) to predict consumption levels. The image patch is fed into the encoder with the classifier head, which outputs the predicted class ($\hat{y_i}$), and is trained with $\mathcal{L}_{task}$.  Data augmentations (e.g vertical and horizontal image flips, 90 degree random rotations and 15\% zooms) are performed during training.
\subsubsection{Model D: Building Characteristics Only} The goal of \textit{Model D} is to provide additional interpretation around the black box CNN in \textit{Model C}. Rather than evaluate the whole image, this model aims to evaluate the importance of \textbf{only} roof characteristics of the building of interest, while ignoring the surrounding context of the household. To achieve this, \textit{Model D} utilizes only building roof characteristics (area and type) as predictors of consumption levels. Specifically, building roof area and the RGB 3-channel intensities are extracted and used as features for prediction. Building roof area and color are inputted into the previously defined MLP to predict consumption levels (Supplementary Materials: A.1.). The MLP is also trained with $\mathcal{L}_{task}$.

\textbf{Roof-top area extraction}: The point indicator approach proposed by \cite{fobi2020learning} is chosen over conventional segmentation because the building polygons available for segmentation (Microsoft Building Footprints in Uganda \cite{microsoftug}) suffer from misaligned and omitted labels when compared to our satellite images. First we select only polygons that are well aligned with structures in satellite imagery. The well-aligned polygons together with the Pointer Segmentation Network\cite{fobi2020learning} are used to train a segmentation model that learns when some of the instances within the images are omitted. This segmentation model was also trained with the dissimilarity loss ($\mathcal{L}_{seg}$). After training on Uganda, we also generate 1000 hand-labelled footprints in Kenya and use the small sample from Kenya to tune the Pointer Segmentation Network. Once the model is tuned to Kenya, the GPS locations of the buildings ($b_i$) in our dataset, are used to obtain a point within each image ($x_i$). This point when combined with the tuned Pointer Segmentation Network is used to extract the footprint for building $b_i$. The extracted footprint is then used to crop out the pixel intensities of the roof. We assume building roofs have a uniform color, thus the roof pixel mean for each channel is used in addition to the roof area (obtained from images pre-electrification) as input features to predict consumption levels after electrification.
\section{Data}
The dataset used in this work has 3 components: 1) Monthly post-paid electricity bills, 2) Overhead daytime satellite imagery, and 3) Public data sources. We unify these 3 data sources by matching the billing dataset to images or public data sources using customer locations within the billing dataset. Following are some details about each. 
\subsection{Ground truth electricity data}
\begin{figure}[t] 
  \centering
  \includegraphics[width=\linewidth]{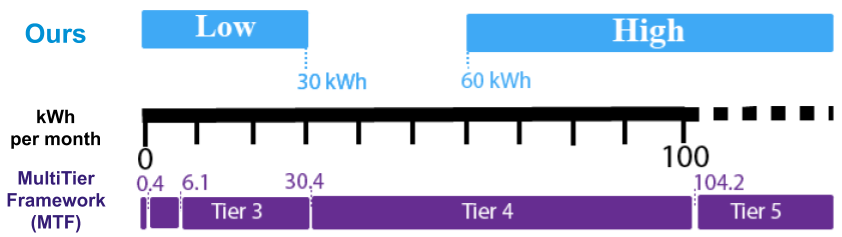}
  \caption{Illustration of World Bank Multi-Tier Framework Consumption Tiers relative to our levels of consumption .}
  \label{eguide_boundaries}
  \vspace{-0.15in}
\end{figure}

Previously\cite{fobi2018longitudinal}, we conducted a longitudinal study of 100k+ randomly sampled electrified households, observing that median customers in Kenya typically reach a consistent level of electricity consumption roughly 12 months after receiving an electricity connection. Given this observation, we define the average monthly consumption of a household after 12 months of a connection as the \textit{expected stable electricity consumption}. For each household, all bills after one year of connection are averaged to obtained a single stable estimate of electricity consumption. The World Bank's Multi-Tier Framework (MTF) divides electricity consumption into a series of Tiers, based on levels of electricity services. We consider low levels of consumption as corresponding to Tiers 0 - 2 of the framework while high consumption levels correspond to $>=$ Tiers 3. Our levels of consumption are obtained by placing a threshold (\textit{thres}) at 30 kWh/month. Figure \ref{eguide_boundaries} illustrates our levels of consumption relative to the MTF tiers. We select a 30 kWh/month boundary because it aligns with MTF break points and energy access practitioners rely on the MTF tiers to support spatial electrification planning.  
Rather than defining the binary class with low being $<=$ 30 kWh and high being $>$ 30 kWh, we select a discontinuous boundary where stable monthly consumptions (kWh) $<=$30 kWh are considered low while $>=$ 60 kWh are considered high \footnote{Customers between 30-60 kWh represent a harder set to study given that we use proxy measures (building characteristics) from satellite imagery. The disjoint boundaries enable electricity planners to still identify customers with low electricity consumption ($<=$30 kWh) to target lower-cost electricity technologies and customers, while also enabling planners to target customers likely to have high electricity consumption ($>=$60 kWh) for more traditional grid-based connections. This design choice, made with significant input from electricity system planning practitioners, supports the twin goals of enhancing the financial sustainability of electricity providers and preserving model performance for the relevant task.}.
To develop a matched dataset of bills and images, customers are grouped by location to obtain electrified buildings. We select residential buildings with only one customer account and these buildings are matched to contemporaneous daytime satellite imagery. From 135,702 Kenya Power customers, 52,083 single customer buildings are used to calculate the monthly stable consumption of each customer. Single customer buildings are chosen over multi-customer buildings because our billing dataset does not contain all the customers in each multi-customer building. Keeping in mind our goal of predicting expected levels of consumption upon electrification ($\hat{y_i}$) using images pre-electrification ($x_i$), satellite image acquisition dates are used to select buildings with satellite imagery acquired prior to the stable consumption phase. Our selection assumes that the socioeconomic benefits of electrification do not become apparent within a daytime satellite image immediately ($<$ 1 year) after the household is electrified. Labels are obtained by applying the discontinuous threshold, to obtain binary consumption levels.
\subsection{Satellite Imagery}
Satellite imagery used in this work consists of 3-band (RGB) 50 cm daytime DigitalGlobe Satellite Imagery obtained between 2002 and 2020. The DigitalGlobe imagery while providing country-wide coverage only contains a single image per location (there are no temporal images for the same location). To train the building segmentation task, images with corresponding building polygons were used irrespective of the image acquisition date. To predict electricity consumption levels, buildings whose images ($x_i$) occurred pre-electrification are selected as part of the training, validation and test datasets. 

\subsection{Public data sources}
\textit{Census Information}: The 2009 Kenya census \cite{2009Kenyacensus} provides low-resolution demographic information on households at the ward administrative level, for which there are 1450 wards in Kenya. The 2009 census is selected over the more recent 2019 census because the recent census data are not yet publicly available and also occur significantly after our electricity consumption data. In addition, the 2009 census better aligns with our formulation for latent electricity prediction using data the occurs prior to when the household was electrified. Table \ref{other_data} shows a summary of parameters obtained from the 2009 Kenya census, grouped by semantic meaning. The census reports the \% of households in a ward for every category. Seventeen census indicators were used as additional data. Customers in the same ward are assigned the ward census value.
\begin{table}[t]
\caption{Non-visual data used for electricity prediction.}
\resizebox{\columnwidth}{!}{%
\begin{tabular}{|l|}\hline
\textbf{Census (\% of ward)}                        \\ \hline
Water Source (Surface | Improved | Unimproved)               \\ \hline
Sanitation (Improved | Unimproved)                  \\ \hline
Lightfuel (Finished | Rudimentary)                  \\ \hline
Floor material (Finished | Rudimentary)             \\ \hline
Cook fuel (Finished | Rudimentary)                  \\ \hline
Wall material (Finished | Rudimentary | Natural)    \\ \hline
Rooftop material (Finished | Rudimentary | Natural) \\ \hline
\textbf{Intensity}                 \\ \hline
VIIRS Nighttime lights         \\ \hline
\end{tabular}}
\label{other_data}
\vspace{-0.15 in}
\end{table}
\textit{Intensity}: 15 arcseconds/pixel (450m at the equator) VIIRS Satellite Nighttime Light data~\cite{viirs} is often used to study economic development and electricity. Average monthly nighttime light intensities for every year (2012 - 2015) were calculated using monthly VIIRS composites. The nighttime light intensity for the year prior to when the building was electrified is retrieved, for the grid cell in which the building is located. If the building was electrified before 2012, the 2012 intensity is used, as VIIRS composites are only available after 2011. 
\section{Experiments \& Results}
\label{results}
\subsection{Experimental Setup}
After matching satellite images pre-electrification to the mean electricity consumption of households in the stable phase, the datasets consist of 20,000 individual households. 75 \% was used for training, 15 \% for validation and 10 \% were held-out as the test set. The distribution of the overall electricity data is preserved within each sub-group of train, val, and test.  Results are reported for the 10 \% in the held-out test set. 
All models were trained with an Adam optimizer with a learning rate of 1e$^{-5}$. This learning rate was chosen over others (1e$^{-3}$, 1e$^{-4}$ and 1e$^{-6}$) as it offered the best overall performance and training convergence. The building segmentation model in Phase 1 was trained for 30 epochs (as both the train and validation curve had converged). The MLP models are trained for 20 epochs and the CNN model is trained for 100 epochs. A batch size of 64 was used and 25\% dropout was applied on all models to prevent overfitting. Feature standardization and normalization was performed. We used an input patch size of 128 x 128 pixels to provide enough field of view that captures the building in the centre and some context around it. 
\subsection{Performance Evaluation}
Table \ref{perfresults} shows the performance of each of the four models presented in Section \ref{models}. Our evaluation metrics include: 1) Class Accuracies shown as True Negative (TN - low consumers) and True Positive (TP - high consumers) 2) Equally weighted F1-score, and 3) Area-Under-Curve (AUC).
\subsubsection{(A) Average Historical Consumption}
Using average historical consumption levels as predictors for yet-to-be connected customers results in a highly-skewed prediction (0.35 F1-Score), with 99\% of high consumers correctly predicted while only 2\% of low consumers are correctly predicted. The strong performance skew is because of the electrification bias, where high consumers (who are often wealthier) are electrified first while lower consumers are added over time. The average historical consumption will always over estimate the consumption levels of the newly connected (often lower consuming) customers. An energy planner using administrative level averages will spend large investments to connect low consumers via grid when cost-effective alternatives might be more suitable. 
\subsubsection{\textbf{(B) Non-Visual Data}}
Census indicators offer a range in F1-scores (0.57 - 0.65), with the highest obtained from rooftop materials. This suggests that building characteristics are important proxy features for predicting consumption levels upon electrification. Census parameters while performing better than \textit{Model A}, also show a performance skew towards the lower consumption class. Nighttime lights only achieved a 0.51 F1-score in predicting individual consumption levels of future electricity connections. Overall F1-scores and AUCs with non-visual is still below that obtained with images.
\begin{table}[t]
\caption{Comparison of electricity prediction models in Kenya. Area-Under-Curve (AUC) \& Balanced F1-score metrics are presented. True Negative (TN) shows the fraction of low consumers that were correctly predicted while True Positive (TP) shows the fraction of high consumers that were correctly predicted.}
\label{perfresults}
\resizebox{\columnwidth}{!}{%
\begin{tabular}{|c|c|c|c|c|c|c|}
\hline
\textbf{Model} & \textbf{Method} & \textbf{Data Input} & \textbf{AUC} & \textbf{F1-score} & \textbf{TN} & \textbf{TP} \\ \hline
A & Historical Consumption & Average kWh & NA & 0.35 & 0.02 & 0.99 \\ \hline
\multirow{7}{*}{B} & \multirow{7}{*}{Census} & i) Water Src. & 0.69 & 0.63 & 0.82 & 0.47 \\ \cline{3-7} 
 &  & ii) Sanitation & 0.62 & 0.57 & 0.63 & 0.51 \\ \cline{3-7} 
 &  & iii) Lighting Fuel & 0.68 & 0.63 & 0.82 & 0.47 \\ \cline{3-7} 
 &  & iv) Floor Mat. & 0.67 & 0.61 & 0.84 & 0.41 \\ \cline{3-7} 
 &  & v) Cooking Fuel & 0.67 & 0.60 & 0.86 & 0.39 \\ \cline{3-7} 
 &  & vi) Wall Mat. & 0.66 & 0.63 & 0.69 & 0.57 \\ \cline{3-7} 
 &  & vii) Rooftop Mat. & 0.69 & 0.65 & 0.66 & 0.64 \\ \hline
B & Nighttime Lights & VIIRS & 0.52 & 0.51 & 0.77 & 0.30 \\ \hline
\textbf{\begin{tabular}[c]{@{}c@{}}C (Ours)\\ Building Seg. Weights\\ without Contrastive Loss\end{tabular}} & \textbf{\begin{tabular}[c]{@{}c@{}}Building Characteristics \\ \& Context\end{tabular}} & \textbf{Images} & \textbf{0.75} & \textbf{0.68} & \textbf{0.70} & \textbf{0.66} \\ \hline
\begin{tabular}[c]{@{}c@{}}C (Ours)\\ Building Seg. Weights\\ with Contrastive Loss\end{tabular} & \begin{tabular}[c]{@{}c@{}}Building Characteristics\\ \& Context\end{tabular} & Images & 0.73 & 0.67 & 0.66 & 0.67 \\ \hline
\multirow{3}{*}{D} & \multirow{3}{*}{Building Characteristics} & Roof Area & 0.65 & 0.61 & 0.66 & 0.56 \\ \cline{3-7} 
 &  & Roof Color & 0.66 & 0.62 & 0.56 & 0.68 \\ \cline{3-7} 
 &  & Roof Area \& Color & 0.69 & 0.64 & 0.65 & 0.64 \\ \hline
B \& C & \begin{tabular}[c]{@{}c@{}}Building Characteristics\\ \& Context,\\ Census,\\ Nighttime Lights\end{tabular} & \begin{tabular}[c]{@{}c@{}}Images,\\ i-vii,\\ VIIRS\end{tabular} & 0.77 & 0.70 & 0.76 & 0.65 \\ \hline
\end{tabular}}
\vspace{-0.1 in}
\end{table}

\subsubsection{\textbf{(C) Building Characteristics and Context}}
Using only daytime satellite images as the basis for prediction, our approach (\textit{Model C}) achieves an balanced F1-score of 0.68 with an AUC of 0.75. This image-based model ensures good performance in both classes (70 \% and 66 \% correctly predicted as low and high respectively). Good performance in both classes is crucial for energy planners (especially in highly heterogeneous regions) and suggests that images better support local class differentiation, compared to the other lower resolution data sources. Our CNN architecture performs comparable to well-known architectures such as VGG16\cite{simonyan2014very} and ResNet50\cite{he2016deep}, even though our custom encoder only has 728k trainable parameters compared to millions in VGG-16 and ResNet-50 (Table \ref{tab:encoderp}).

\begin{table}[b]
\centering
\vspace{-0.1 in}
\caption{Performance comparison of well-known architectures compared to our encoder}
\label{tab:encoderp}
\begin{tabular}{|c|l|c|l|}
\hline
 & Weights & F1-score & \# Parameters \\ \hline
VGG16 & Random & 0.62 & 14,714,688 \\ \hline
Resnet-50 & Random & 0.62 & 23,587,712 \\ \hline
Our Encoder & Random & 0.63 & 728,0065 \\ \hline
\end{tabular}
\end{table}
The classifier encoder was pretrained on a building segmentation task. The value of this pretraining step is validated by 2 approaches. First, we compare classification performance with and without pretraining and noticed that pretraining the encoder on building segmentation improves the electrification classifier accuracy from 0.63 to 0.68 when all the electricity training data is used. When the training data is reduced, performance is preserved for the building segmentation-pretrained model when compared to a model trained from scratch (Supplementary Materials: A.2.). Second, we also compare the performance with and without a supervised constrastive loss\cite{khosla2020supervised} when initialized with building segmentation weights. Here, we hypothesize that if relevant embeddings are obtained through building segmentation, further optimizations of the embeddings (through a contrastive loss) would provide no additional performance gains. The classifier (initialized with building segmentation weights) was trained with a supervised contrastive loss (temperatures: 0.08 and 0.1) prior to finetuning the final layer for classification. Adding a contrastive loss did not further improve performance (0.67 F1-score), suggesting that the building segmentation task learnt relevant embeddings needed for the classification task.

Combining visual and non-visual features (\textit{Model B \& C}) through a multi-modal architecture (Supplementary Materials: A.3.), increased the F1-score to 0.70. Using multi-modal data can be helpful to improve electricity predictions. 

The image model was evaluated on households with monthly consumption between 31-59 kWh. We observed a 4\% decline in F1-score when a threshold of $<=$30 kWh and $>$ 30 kWh is used for low and high, respectively.
\footnote{This set make up 28 \% of single household customers within our data.}
\subsubsection{(D): Building Characteristics Only}
\begin{figure}[t]
  \vspace{-0.1 in}
 \centering
  \includegraphics[width=0.9\linewidth]{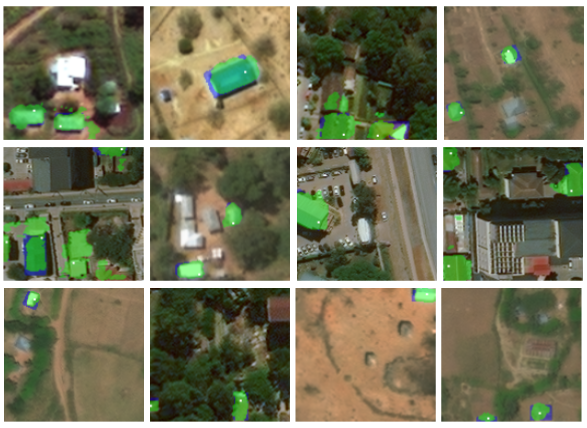}
 \caption{Sample segmentation outputs using an indicator point to specify which building(s) to segment\cite{fobi2020learning}. White dots show input points given to the model to specify which buildings to segment. \textcolor{green}{Green} shows predictions and \textcolor{blue}{blue} ground-truth.}
 \label{sample_segmentation}
 \vspace{-0.1 in}
\end{figure}

Complementary to the CNN, \textit{Model D} isolates and quantifies the relevance of building characteristics (\textbf{only roof-top area and type}) when learning to predict consumption. Prior to discussing \textit{Model D}'s performance, we first discuss the performance of the pointer segmentation model used to obtain building footprints.
\textbf{Performance of rooftop segmentation:}
Hand-labelled polygons in Kenya showed a validation Intersection-Over-Union (IOU) of \textbf{0.54}.
Figure \ref{sample_segmentation} shows sample segmented footprints in Kenya given indicator points (white dots) specifying buildings. This segmentation model was applied on the electricity training, validation and test set to extract building footprints (roof area) and average roof pixel intensities for each channel (rooftop type). 

\textbf{Model D Performance:}
Roof area and roof color features respectively predicted 66\% and 56\% of the low class correctly while respectively predicting 56\% and 68\% of the high class correctly. However, combining both roof area and color reduced the skew in performance while encouraging better predictions for both low and high levels of consumption. This suggests that individual roof sizes may be more indicative of low consumers while roof materials (from mean pixel intensities) are helpful for better identifying high consumers. 
Direct use of images, which includes both the building characteristics and the surrounding context of the building improves the F1-score (relative to using only building characteristics) by 4\% . This added benefit is likely a combined effect of bypassing segmentation error and the additional information found within the surrounding context of the building.
\begin{figure}[b]
\vspace{-0.1 in}
  \centering
  \includegraphics[width=1\linewidth]{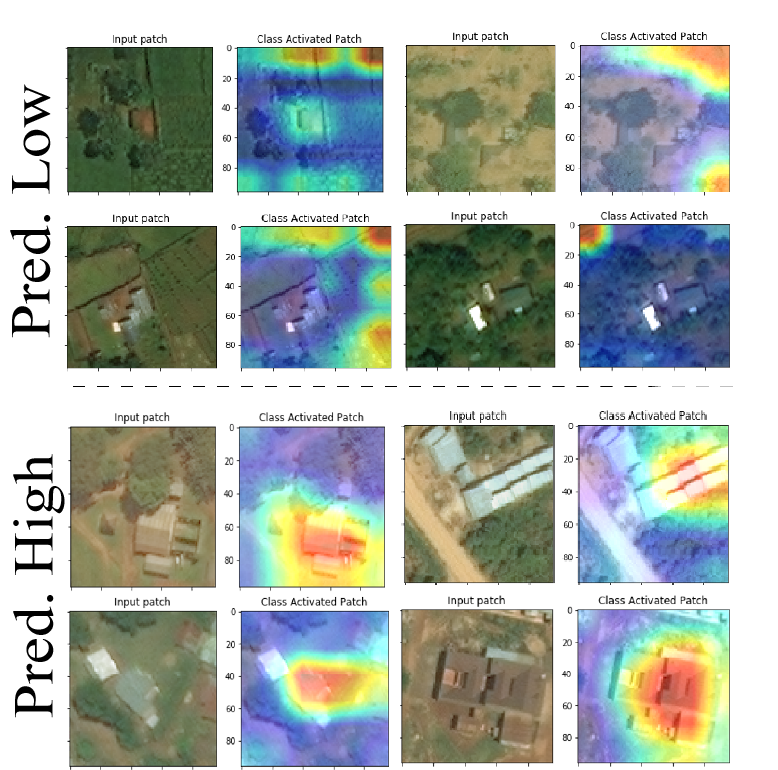}
  \caption{Gradient-based class activation maps for sample in test set. Stronger neural activations are in \textcolor{red}{Red} while weaker neural activations are in \textcolor{blue}{Blue}. Buildings are strongly activated when predicting high levels of consumption while the activation is more distributed between the building and its surrounding context when predicting low levels of consumption.}
  \label{grad_cam}
  \vspace{-0.2 in}
\end{figure}

\subsection{GRAD-CAM (Qualitative Interpretation):}
Class activation maps are used to qualitatively explain the relevant features that the image-based classifier in \textit{Model C} uses to predict levels of electricity consumption. GRAD-CAM \cite{selvaraju2017grad} is used to visualize portions of the image that have high neural activations when making predictions. Some GRAD-CAM visualizations are shown in Figure~\ref{grad_cam}. Strong activations (red) on buildings are observed when predicting high-consuming buildings while the activation is more distributed between the building and its surrounding context (blue) when predicting low-consuming buildings. The image-based model utilizes both building size and surrounding land as indicators of consumption levels.

\subsection{Validation with independent survey data}
We present an extra validation of our approach against an independently collected and nationally representative baseline household survey (4473 households) of both electrified and unelectrified households\cite{mtfdata}. The Kenya Multi-Tier Framework (MTF) Survey conducted between 2016 - 2018, asks grid-connected households how much electricity they consumed in the most recent month. The reported and binned onetime consumptions are compared with our predictions for Kenya Power residential grid-connected customers. Images alone are used to predict consumption levels of $~$5.3 million Kenya Power residential customers connected by the start of 2016. MTF samples are binned at $<=$30 kWh as low and $>=$60 as high. Sample weighted Pearson correlations (Table \ref{mtf_performance}) between MTF and predicted consumption levels are reported for 29 counties with at least 15 MTF survey samples of grid-connected customers. Using all 29 counties, a correlation of 0.64 is observed. Excluding Nairobi county increases, the correlation to 0.82\footnote{Nairobi (the largest city in Kenya), is excluded because the survey over-samples recently-electrified informal settlements.}. These correlations show strong agreement (p $<$0.0005) given an independent source of national data. 
\begin{table}[t]
\caption{County-level consistency between independently collected Multi-Tier Framework (MTF) Survey and predictions for $~5.3$ million Kenya Power residential customers. Results (p-value $<$0.0005) for counties with at least 15 MTF samples.}
\label{mtf_performance}
\centering
\begin{tabular}{|l|l|l|}
\hline
                    &  \begin{tabular}[c]{@{}l@{}} 29 Counties \\ \end{tabular} & \begin{tabular}[c]{@{}l@{}}28 Counties\\  (Excluding Nairobi)\end{tabular} \\ \hline
Pearson correlation & 0.64              & 0.82                                                                       \\ \hline
\end{tabular}
\vspace{-0.1 in}
\end{table}
\section{Conclusion \& Future work}


This paper proposes a method to estimate levels of electricity consumption for unconnected households using pre-electrification images. Our results show that our novel methodology of using satellite images for electricity prediction outperforms existing approaches currently used by energy planners. We also present a multi-modal approach that combines satellite images with other data sources to further improve the overall prediction performance. The predictions of our model (currently deployed in Kenya) provide a birds-eye view of relative levels of consumption upon electrification throughout the country and equip decision-makers with a direct measure of expected energy usage as well as a novel proxy for economic activity. This can enable better system planning and stretch investments in electrification to connect more people to modern energy sources.

Predicting electricity usage from satellite images remains a difficult task, mainly because elements in satellite images (rooftops, roads, fields) are only proxy measures for electricity. Utilities could improve their own predictions with additional much larger and comprehensive data (e.g. bills and locations of all existing customers) that they possess. We are keen to co-develop such methodologies with partners. We also plan to evaluate the transferability of our method by extending our approach to other countries and sectors (e.g., commercial and industrial).


{\small
\bibliographystyle{ieee_fullname}
\bibliography{egbib}
}
\newpage
\appendix 
\section{Supplementary Material}
\subsection{Multi-Layer Perception (MLP) Architecture}
\begin{figure}[b]
  \centering
  \includegraphics[width=\linewidth]{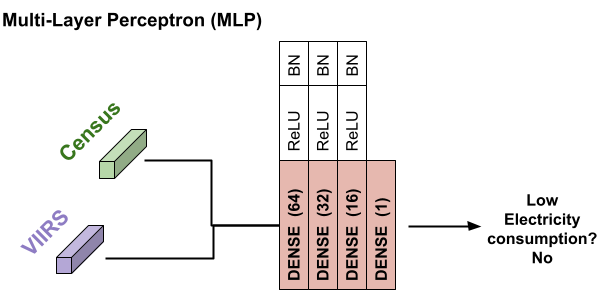}
  \caption{MLP architecture used to train \textit{Model B} and \textit{Model D}}
  \label{mlp}
  \vspace{-0.1 in}
\end{figure}

Figure \ref{mlp} shows the MLP architecture used to train\textit{ Model B} (Non-visual Data) and \textit{Model D} (Building Characteristics Only). The MLP consists of 3 dense layers with 64, 32, and 16 filters respectively, all with ReLU activations. The last dense layer consists of a softmax activation. 25 \% dropout was applied to minimize overfitting.  
\subsection{Performance of building segmentation}
\begin{figure}[t]
  \centering
  \includegraphics[width=\linewidth]{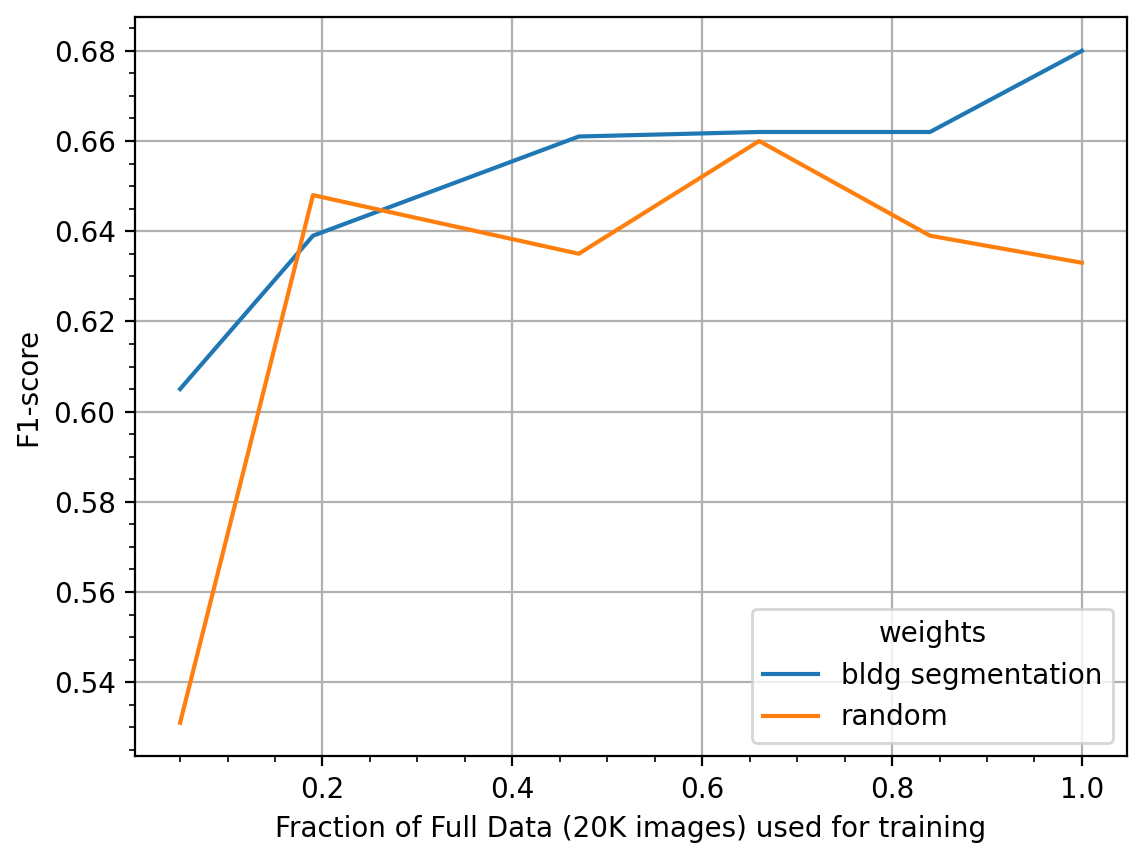}
  \caption{Comparison of prediction performance when the classifier is initialized with random weights versus building segmentation weights. Learning about building segmentation improves performance in low-data regimes and makes performance less susceptible to harder labels thereby offering a regularizing effect.}
  \label{pretraining}
  \vspace{-0.2 in}
\end{figure}

Additional evaluation of the building segmentation task is done by observing how the classifier performs at varying training data sample sizes. Figure \ref{pretraining} shows the F1-score at different training data sample sizes when random subsets of the data are selected and either random weights or building segmentation weights are used to initialize model training. 
At each sample size increment, samples from the previous sample size are included. E.g. the 20 \% dataset contains all the samples from the 5 \% dataset. 
Initializing with building segmentation weights offers performance gains especially at smaller sample sizes. The improved performances with building segmentation weights suggests that underlying characteristics about buildings (rooftop type, color, size) provides relevant features for consumption prediction. This is inline with our initial findings that building characteristics are relevant in predicting consumption levels. In addition to improved model performance, building segmentation weights make the classifier less susceptible to label quality. Specifically when random weights are used for initialization, it is observed that the randomly selected sub-sample at 60 \% of the full dataset, performed the best and performance dropped as more samples were added. This suggests that the ease $|$ difficulty of the sub-sample significantly affects performance. Building segmentation weights initializes the model in a suitable learning space and has a regularizing effect even as harder labels may be introduced, allowing only additional useful information to be extracted.
Obtaining large amounts of useful samples to appropriately predict consumption of yet to be connected customers  can be challenging. For energy practitioner looking to apply our approach, we show that learning about buildings from using a segmentation task, provides useful weight tuning needed for appropriate prediction of consumption tiers.
\subsection{Multi-modal architecture: Encoder and MLP}
\begin{figure}[h]
  \centering
  \includegraphics[width=1.1\linewidth]{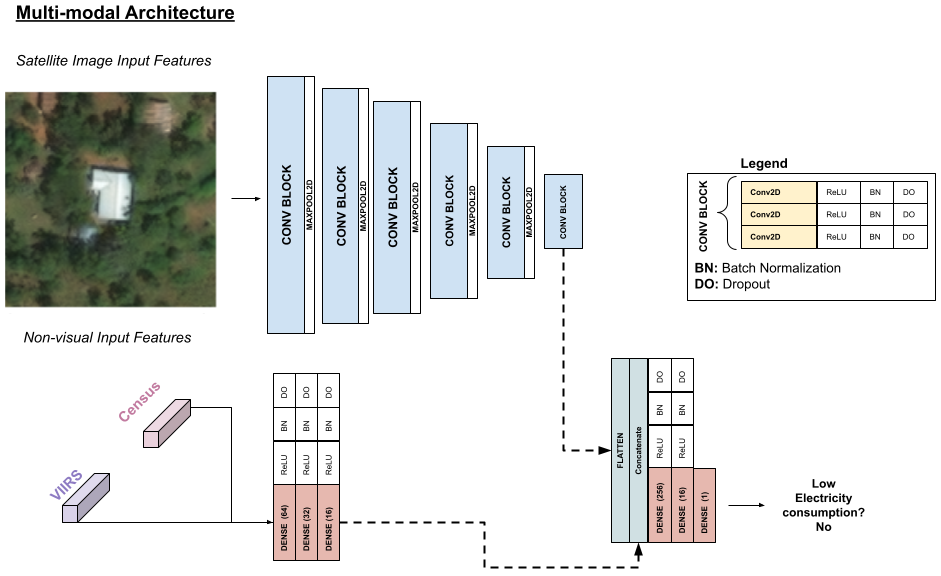}
  \caption{Multi-modal architecture combining the CNN image-based encoder with an MLP to predict consumption levels using visual images and non-visual public data sources.}
  \label{multimodal_arch}
\end{figure}

Figure \ref{multimodal_arch} shows the multimodal architecture used to combine satellite images with public data sources. 

\end{document}